\documentclass{applemlr}
\usepackage{amsmath}
\usepackage{enumerate} 
\usepackage{algorithm}
\usepackage{algpseudocode}
\usepackage{amsfonts}
\usepackage{amsthm}
\usepackage{newtxtt}
\usepackage{cleveref}
\usepackage{diagbox} 
\usepackage{colortbl}
\usepackage{amssymb}
\usepackage{xspace}
\usepackage{wrapfig}
\usepackage{adjustbox}
\usepackage{tabularx}
\usepackage{booktabs}
\usepackage{mathtools}
\usepackage{tikz}
\usepackage{enumitem}
\usepackage{silence}
\usepackage{dsfont}
\usepackage[table]{xcolor}
\usepackage[dvipsnames]{xcolor}
\usepackage{multirow}
\usepackage{makecell}
\usepackage{xfakebold}

\usepackage{amsmath,amsfonts,bm}

\def\eqref#1{equation~\ref{#1}}

\def\1{\bm{1}}

\DeclareMathAlphabet{\mathsfit}{\encodingdefault}{\sfdefault}{m}{sl}
\SetMathAlphabet{\mathsfit}{bold}{\encodingdefault}{\sfdefault}{bx}{n}

\usepackage{multirow}
\usepackage[normalem]{ulem}
\useunder{\uline}{\ul}{}
\definecolor{AppleWhite}{RGB}{255,255,255}
\definecolor{ApplePrimaryCoolGray}{RGB}{116,128,139}
\definecolor{AppleCoolGray1}{RGB}{199,209,214}
\definecolor{AppleCoolGray2}{RGB}{147,174,190}
\definecolor{AppleCoolGray3}{RGB}{124,147,160}
\definecolor{AppleCoolGray4}{RGB}{92,102,109}
\definecolor{AppleCoolGray5}{RGB}{78,93,100}
\definecolor{AppleCoolGray6}{RGB}{53,60,65}
\definecolor{AppleBlack}{RGB}{0,0,0}
\definecolor{AppleSecondaryChartGray}{RGB}{168,168,168}
\definecolor{AppleChartGray2}{RGB}{233,233,233}
\definecolor{AppleChartGray3}{RGB}{211,211,211}
\definecolor{AppleChartGray4}{RGB}{190,190,190}
\definecolor{AppleChartGray5}{RGB}{140,140,140}
\definecolor{AppleChartGray6}{RGB}{102,102,102}
\definecolor{AppleChartGray7}{RGB}{64,64,64}
\definecolor{ApplePrimaryChartBlue}{RGB}{84,151,193}
\definecolor{AppleBlue2}{RGB}{212,229,239}
\definecolor{AppleBlue3}{RGB}{169,202,223}
\definecolor{AppleBlue4}{RGB}{127,177,209}
\definecolor{AppleBlue5}{RGB}{71,130,166}
\definecolor{AppleBlue6}{RGB}{55,99,128}
\definecolor{AppleBlue7}{RGB}{45,72,89}
\definecolor{ApplePrimaryChartGreen}{RGB}{83,172,121}
\definecolor{AppleGreen2}{RGB}{212,234,221}
\definecolor{AppleGreen3}{RGB}{169,213,188}
\definecolor{AppleGreen4}{RGB}{126,193,155}
\definecolor{AppleGreen5}{RGB}{58,140,82}
\definecolor{AppleGreen6}{RGB}{39,102,54}
\definecolor{AppleGreen7}{RGB}{29,58,31}
\definecolor{ApplePrimaryChartYellow}{RGB}{253,195,93}
\definecolor{AppleYellow2}{RGB}{254,240,214}
\definecolor{AppleYellow3}{RGB}{254,224,174}
\definecolor{AppleYellow4}{RGB}{254,210,134}
\definecolor{AppleYellow5}{RGB}{230,168,69}
\definecolor{AppleYellow6}{RGB}{191,131,46}
\definecolor{AppleYellow7}{RGB}{153,107,54}
\definecolor{ApplePrimaryChartOrange}{RGB}{250,151,92}
\definecolor{AppleOrange2}{RGB}{254,229,214}
\definecolor{AppleOrange3}{RGB}{252,203,173}
\definecolor{AppleOrange4}{RGB}{252,178,133}
\definecolor{AppleOrange5}{RGB}{227,121,68}
\definecolor{AppleOrange6}{RGB}{191,87,46}
\definecolor{AppleOrange7}{RGB}{143,59,36}
\definecolor{ApplePrimaryChartRed}{RGB}{227,94,105}
\definecolor{AppleRed2}{RGB}{248,215,217}
\definecolor{AppleRed3}{RGB}{241,174,180}
\definecolor{AppleRed4}{RGB}{234,135,143}
\definecolor{AppleRed5}{RGB}{196,63,77}
\definecolor{AppleRed6}{RGB}{153,35,53}
\definecolor{AppleRed7}{RGB}{102,19,43}
\definecolor{ApplePrimaryChartPurple}{RGB}{161,150,204}
\definecolor{ApplePurple2}{RGB}{231,228,242}
\definecolor{ApplePurple3}{RGB}{208,202,229}
\definecolor{ApplePurple4}{RGB}{185,176,217}
\definecolor{ApplePurple5}{RGB}{128,113,171}
\definecolor{ApplePurple6}{RGB}{89,76,128}
\definecolor{ApplePurple7}{RGB}{62,46,101}
\definecolor{AppleCoolGray}{RGB}{116,128,139}
\definecolor{AppleChartGray}{RGB}{168,168,168}
\definecolor{AppleBlue}{RGB}{84,151,193}
\definecolor{AppleGreen}{RGB}{83,172,121}
\definecolor{AppleYellow}{RGB}{253,195,93}
\definecolor{AppleOrange}{RGB}{250,151,92}
\definecolor{AppleRed}{RGB}{227,94,105}
\definecolor{ApplePurple}{RGB}{161,150,204}
\usepackage[table]{xcolor}
\usepackage{wrapfig}

\definecolor{textgray}{HTML}{6E6E73}
\usetikzlibrary{positioning, calc}
\usetikzlibrary{decorations.pathmorphing}

\makeatletter
\patchcmd{\wrong@fontshape}{\@gobbletwo}{}{}{}
\makeatother
\numberwithin{equation}{section} 
\tcbuselibrary{minted}
\setminted[python]{
  linenos,
  breaklines,
  fontsize=\footnotesize,
  xleftmargin=2em
}

\newcommand{\ours}{RL4HS\xspace}

\makeatletter
\AtBeginDocument{
  \urlstyle{sf}
  
}
\makeatother

\definecolor{light}{RGB}{125, 125, 125}
\crefname{tcb@cnt@pbox}{code}{code}
\Crefname{tcb@cnt@pbox}{Code}{Code}
\crefname{assumption}{assumption}{assumption}
\Crefname{assumption}{Assumption}{Assumptions}

\newtcolorbox[auto counter]{pbox}[2][]{
  colback=white,
  title=Code~\thetcbcounter: #2,
  #1,fonttitle=\sffamily,
  fontupper=\sffamily,
  arc=2pt,
  colframe=bgcolor,
  coltitle=fgcolor,
  colbacktitle=bgcolor,
  toptitle=0.25cm,
  bottomtitle=0.125cm
}

\makeatletter
\newcommand\applefootnote[1]{%
  \begingroup
  \renewcommand\thefootnote{}%
  \renewcommand\@makefntext[1]{\noindent##1}%
  \footnote{#1}%
  \addtocounter{footnote}{-1}%
  \endgroup
}
\makeatother

\definecolor{cverbbg}{gray}{0.90}

\usepackage[stable]{footmisc}

\title{Learning to Reason for Hallucination Span Detection}
\author{
Hsuan Su$^\heartsuit$$^\diamondsuit$$^*$\qquad
    Ting-Yao Hu$^\diamondsuit$ \qquad
    Hema Swetha Koppula$^\diamondsuit$ \qquad
    Kundan Krishna$^\diamondsuit$ \qquad
    Hadi Pouransari$^\diamondsuit$ \qquad \\
    \textbf{Cheng-Yu Hsieh$^\diamondsuit$} \qquad
    \textbf{Cem Koc$^\diamondsuit$} \qquad 
    \textbf{Joseph Yitan Cheng$^\diamondsuit$} \qquad
    \textbf{Oncel Tuzel$^\diamondsuit$} \qquad 
    \textbf{Raviteja Vemulapalli$^\diamondsuit$}
}
\affiliation{$^\heartsuit$National Taiwan University \qquad $^\diamondsuit$Apple}

\metadata[Correspondence]{\sffamily Ting-Yao Hu: \url{tingyao_hu@apple.com}; Raviteja Vemulapalli: \url{r_vemulapalli@apple.com} }
\date{\sffamily\today}
\abstract{
Large language models (LLMs) often generate hallucinations---unsupported content that undermines reliability. While most prior works frame hallucination detection as a binary task, many real-world applications require identifying hallucinated spans, which is a multi-step decision making process.
This naturally raises the question of whether explicit reasoning can help the complex task of detecting hallucination spans. 
To answer this question, we first evaluate pretrained models with and without Chain-of-Thought (CoT) reasoning, and show that  CoT reasoning has the potential to generate at least one correct answer when sampled multiple times. Motivated by this, we propose RL4HS, a reinforcement learning framework that incentivizes reasoning with a span-level reward function. 
RL4HS builds on Group Relative Policy Optimization and introduces Class-Aware Policy Optimization to mitigate reward imbalance issue. Experiments on the RAGTruth benchmark (summarization, question answering, data-to-text) show that RL4HS surpasses pretrained reasoning models and supervised fine-tuning, demonstrating the necessity of reinforcement learning with span-level rewards for detecting hallucination spans.}

\begin{document}
\maketitle
\begingroup
\renewcommand\thefootnote{}\footnotetext{*~Work done during an internship at Apple.}
\endgroup
\section{Introduction}

Over the past few years, Large Language Models (LLMs) have demonstrated remarkable capabilities across a wide range of natural language processing tasks \citep{xie2023empirical, zhang2023sentimentanalysiseralarge,gao-etal-2024-self,openai2024gpt4technicalreport}.
However, they are still prone to generating \textit{hallucinations}---content that is not supported by the input context or the underlying knowledge sources \citep{zhu2024haluevalwildevaluatinghallucinationslanguage,kalai2025languagemodelshallucinate, Huang_2025}. Hallucinations pose critical risks in downstream applications such as summarization and long-form question answering, where reliability and factual consistency with respect to the input context are paramount. Hence, the ability to detect hallucinations is crucial for successful real-world deployment of LLMs.

Most existing research works focus on \textit{binary hallucination detection} problem, where the goal is to determine if the model output contains hallucinations or not \citep{yang-etal-2024-reassess, yang-etal-2024-fizz, tang-etal-2024-minicheck, ravi2024lynxopensourcehallucination, ji2024llminternalstatesreveal, chuang-etal-2024-lookback}. 
While useful, this formulation is limited: in many real-world applications, one often needs to know which specific spans in the model output are hallucinated in order to assess the reliability of the generated content. 
This motivates the problem of \textit{hallucination span detection}, where the goal is to precisely locate unsupported content in the model output \citep{wu2023ragtruth,  ogasa2025hallucinatedspandetectionmultiview}.

Recently, \textit{reasoning}---the process of systematically arriving at conclusions by generating and utilizing intermediate steps---has been shown to significantly enhance the capabilities of LLMs in solving complex tasks such as mathematics \citep{shao2024deepseekmathpushinglimitsmathematical, yu2025dapoopensourcellmreinforcement} and coding \citep{code-r1, chen2025r1codeinterpretertrainingllmsreason}.
Hallucination span detection is also a complex multi-step decision making process as it requires carefully analyzing the model output to extract all the stated facts and verifying whether each of these facts is fully supported by the input context, and could benefit significantly from a learned reasoning process.

Some existing hallucination detection works \citep{luo2023chatgptfactualinconsistencyevaluator,eliav2025clattercomprehensiveentailmentreasoning} explored Chain-of-Thought (CoT) prompting, and showed that simple CoT can lead to considerable improvements in binary hallucination detection performance providing motivating evidence to explore reasoning for hallucination detection. However, these works do not focus on the fine-grained hallucination span detection problem and they do not explore training a reasoning model for hallucination detection. In this work, we focus on concretely answering the following two research questions: (i) Is learned reasoning process helpful for hallucination span detection? How to learn an effective reasoning process for this task? (ii) Is it necessary to learn a reasoning process specifically for hallucination span detection or do existing general-domain reasoning models suffice for this specific task?

To answer the first question, we train a CoT reasoning-based hallucination span detection model using Reinforcement Learning (RL). Specifically, we train the model on a dataset labeled with hallucination spans using Group Relative Policy Optimization (GRPO;  \cite{shao2024deepseekmathpushinglimitsmathematical}) with a reward function based on the target span-F1 metric. To the best of our knowledge, this is the first work training a reasoning-based hallucination span detection model using RL. The resulting model significantly outperforms a non-reasoning model trained for span detection using Supervised Finetuning (SFT) on the same training dataset. This clearly shows that the reasoning process learned using RL is highly beneficial for detecting hallucination spans. 

While the reward based on span-F1 score is effective, we notice that its asymmetric nature over-incentivizes non-hallucination predictions due to the normalization used in GRPO advantage calculation. To address this issue, we propose a modified version of GRPO, which we refer to as class-aware policy optimization, by introducing a scaling factor for the advantages computed for non-hallucination samples. By using a value smaller than one for this scaling factor, we are able to achieve a better balance between hallucination and non-hallucination classes leading to an overall higher span-F1 score.

To answer the second question, we evaluate several recent reasoning models that have been trained with data from various domains such as mathematics, coding, tool-calling, etc.  Our evaluation results show that, despite being much larger in size, state-of-the-art reasoning models perform significantly worse than a 7B reasoning model trained specifically for hallucination span detection. 

\textbf{Major contributions:} (i) We train a hallucination span detection model using reinforcement learning with span-level reward, and show that the resulting reasoning process improves the hallucination span detection performance by a significant margin when compared to a non-reasoning model trained with the same dataset. (ii) We show that existing reasoning models perform significantly worse when compared to a reasoning model specifically trained for hallucination span detection using RL with span-F1 reward. (iii) We identify an issue with span-F1 reward that leads to over-emphasis on non-hallucination predictions in the context of GRPO, and propose class-aware policy optimization to address this issue.


\section{Hallucination Span Detection}
\subsection{Task}
This paper focuses on the task of hallucination span detection in the context of Conditional Natural Language Generation (CNLG) tasks such as summarization and long-form question answering. Given the input context $c$ and the generated response $y=(y_1, y_2...y_T)$ consisting of $T$ characters, the goal is to identify all the hallucinated spans, which are text segments in $y$ that are not supported by $c$. Each hallucinated span $s$ is represented using its start and end positions in $y$.

\subsection{Model}  
Existing works on hallucination span detection train either a decoder-based generative model that directly outputs hallucinated content as a list of text segments \citep{wu2023ragtruth} or an encoder-based discriminative model that performs token-level binary classification \citep{ogasa2025hallucinatedspandetectionmultiview}. While generative models are a natural fit for exploring CoT reasoning, it is unclear how reasoning can be incorporated into token-level binary classifiers. Hence, in this work, we follow the generative modeling approach of  \cite{wu2023ragtruth} and train an LLM to directly output a list of hallucinated text segments. For each predicted text segment, we get the corresponding span start and end index in $y$ by searching for matching content.

\subsection{Evaluation Metric}
For comparing model predictions with groundtruth, we use the dataset-level span-F1 metric defined in \citet{wu2023ragtruth}. Given the groundtruth spans $S = \{s_m = \left[i_m, j_m\right] \}_{m=1}^M$ and the predicted spans $\hat{S} = \{s_n = \left[i_n, j_n\right] \}_{n=1}^N$, the span-F1 metric is computed using
\begin{equation}
\text{F1} = \frac{2 \cdot \text{Precision} \cdot \text{Recall}}{\text{Precision} + \text{Recall}},\quad
\text{Precision} = \frac{|\mathcal{P} \cap \mathcal{G}|}{|\mathcal{P}|}, \quad
\text{Recall} = \frac{|\mathcal{P} \cap \mathcal{G}|}{|\mathcal{G}|},
\end{equation}
where $\mathcal{G} = \bigcup\limits_{m=1}^M s_m$ and $\mathcal{P} = \bigcup\limits_{n=1}^N s_n$. Here, $\cup$ denotes set union, $\cap$ denotes set intersection, $|.|$ denotes set cardinality, and $[i, j]$ denotes the set of integers from $i$ to $j$.



\section{\ours: Reinforcement Learning for Hallucination Span Detection}
\begin{figure*}[tp!]
    \begin{adjustbox}{width=\linewidth}
    \centering
    \includegraphics{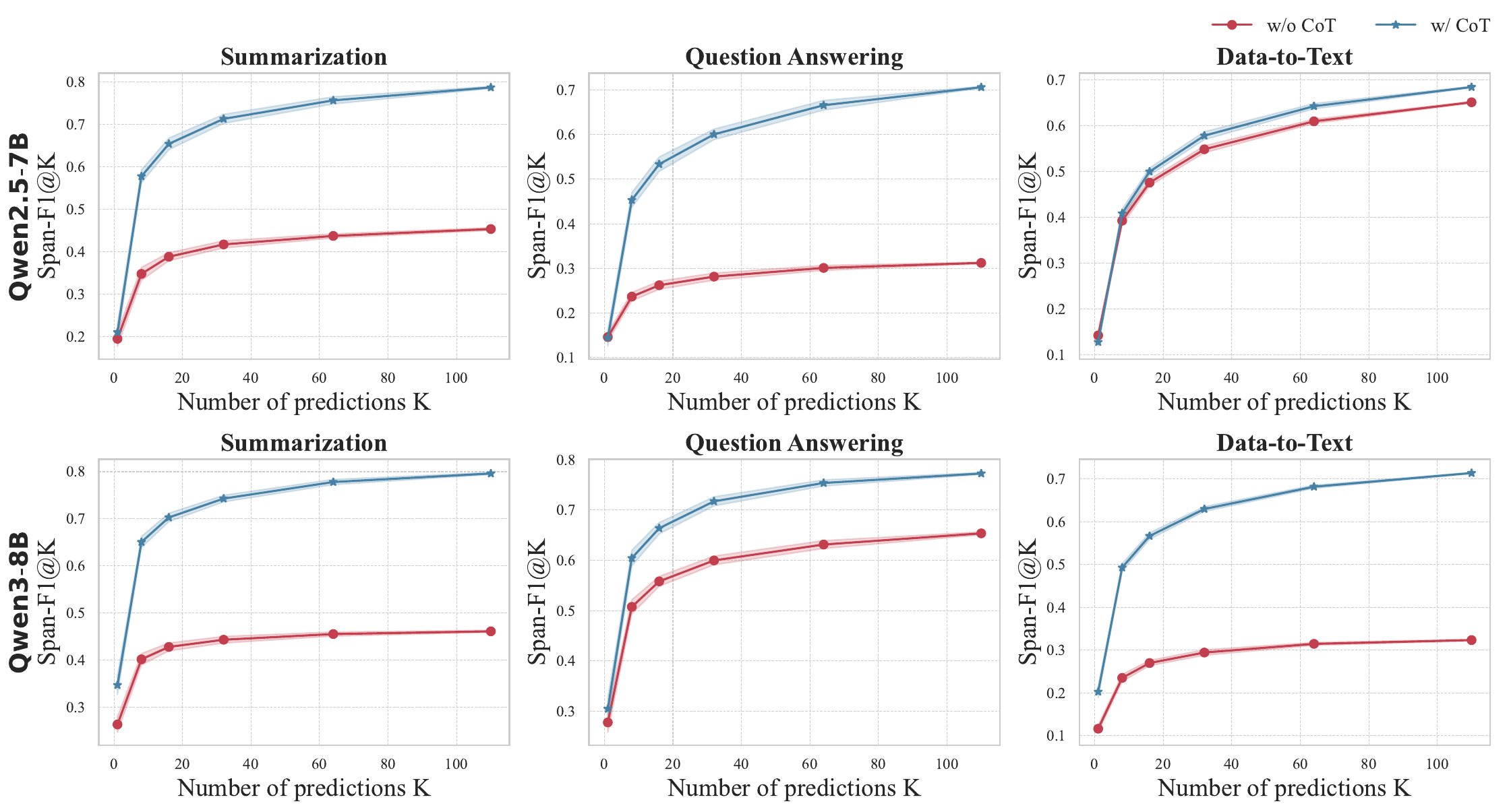}
    \end{adjustbox}
    \caption{\textbf{Span-F1@K for different number of predictions $K$.} Using CoT reasoning provides significant boost as $K$ increases clearly demonstrating the potential of CoT reasoning.}  
     \label{fig:f1@k}
\end{figure*}

\subsection{Motivating RL with diverse CoT reasoning}
\label{sec:cot}
A central question in this study is whether explicit reasoning is beneficial for identifying hallucination spans. As a preliminary experiment, we evaluated \texttt{Qwen2.5-7B}~\footnote{We use the instruct version.}\citep{qwen2.5} and \texttt{Qwen3-8B}~\footnote{We use the reasoning mode and non-reasoning mode with non-COT prompt as elaborated in Qwen3.} \citep{yang2025qwen3technicalreport} models with and without CoT reasoning on data from three CNLG tasks, namely summarization, question answering and data-to-text using the RAGTruth dataset~\citep{wu2023ragtruth}. In CoT reasoning mode, the model is encouraged to first reason about the factual consistency between the input context and the generated output, and then predict hallucinated spans. In the non-reasoning mode, the prompt given to the model instructs it to directly prediction hallucination spans without generating any intermediate content. For each input, the model is run $K$ times and the best prediction is selected based on span-F1. We repeat this experiment for different values of $K$ and show the corresponding Span-F1@K results in Figure~\ref{fig:f1@k}.

At $K=1$, CoT reasoning provides no gains for \texttt{Qwen2.5-7B} and limited gains for \texttt{Qwen3-8B}. However, as $K$ increases, the gap in terms of Span-F1@K increases significantly demonstrating the potential of CoT reasoning to generate at least one accurate prediction when sampled multiple times. These results provide clear motivation to use reinforcement learning for bringing the reasoning capacity of LLMs related to hallucination span detection to the forefront.

We also conducted this experiment with \texttt{Qwen2.5-14B} and \texttt{Qwen3-14B} models and observed a similar behavior. See Appendix~\ref{sec:f1@kapp} for details.

\subsection{RL with GRPO}
As our reinforcement learning framework, we employ Group Relative Policy Optimization (GRPO) \cite{shao2024deepseekmathpushinglimitsmathematical}. 
Unlike Proximal Policy Optimization (PPO) \cite{schulman2017proximalpolicyoptimizationalgorithms}, GRPO eliminates the explicit value function and instead computes baselines from relative group scores. 
The learning objective is defined as:
\begin{equation}
\mathcal{L}_{\text{GRPO}}(\theta) = 
\mathbb{E}_{\tau \sim \pi_\theta} \left[
    \min \left( 
        r_\theta(\tau) A(\tau),\,
        \text{clip}\!\left( r_\theta(\tau), 1 - \epsilon, 1 + \epsilon \right) A(\tau)
    \right)
\right],
\end{equation}

where $\tau$ denotes a trajectory sampled from the current policy $\pi_\theta$, and 
$r_\theta(\tau) = \frac{\pi_\theta(\tau)}{\pi_{\text{old}}(\tau)}$ 
is the probability ratio between the updated and reference policies at each step. 
Instead of relying on a critic network as in PPO, GRPO defines the advantage purely from group-based returns $\{R_i\}_{i \in G(\tau)}$:
\begin{equation}
A(\tau) = \frac{R_\tau - \text{mean}\!\left(\{R_i\}_{i \in G(\tau)}\right)}
{\text{std}\!\left(\{R_i\}_{i \in G(\tau)}\right)}.
\end{equation}

In this formulation, the baseline is determined by the average performance of the group, normalized by its standard deviation, making GRPO particularly suited for scenarios where relative ranking within a group is more informative than absolute value estimates.


\subsubsection{Verifiable Span-F1 Reward}

To apply GRPO for hallucination span detection, we directly use the target span-F1 metric to define the reward. Let $\hat{S}$ be the predicted hallucination spans and $S$ be the ground-truth spans. Then, the reward is defined as 
\[
r_{\text{span}} = 
\begin{cases}
1, & \text{if } \hat{S} = \varnothing \text{ and } S = \varnothing, \\[6pt]
\text{span-F1}(\hat{S}, S), & \text{otherwise}.
\end{cases}
\]
This formulation naturally handles both hallucination and non-hallucination cases. If no hallucinations exist and none are predicted, the model receives maximum reward ($r_{\text{span}}=1$). In other cases, the reward reflects the quality of overlap between predicted and groundtruth spans.

\subsection{Reward Imbalance Across Classes}

\begin{wrapfigure}{r}{0.4\textwidth}
\vspace{-5em}
  \centering
  \includegraphics[width=0.4\textwidth]{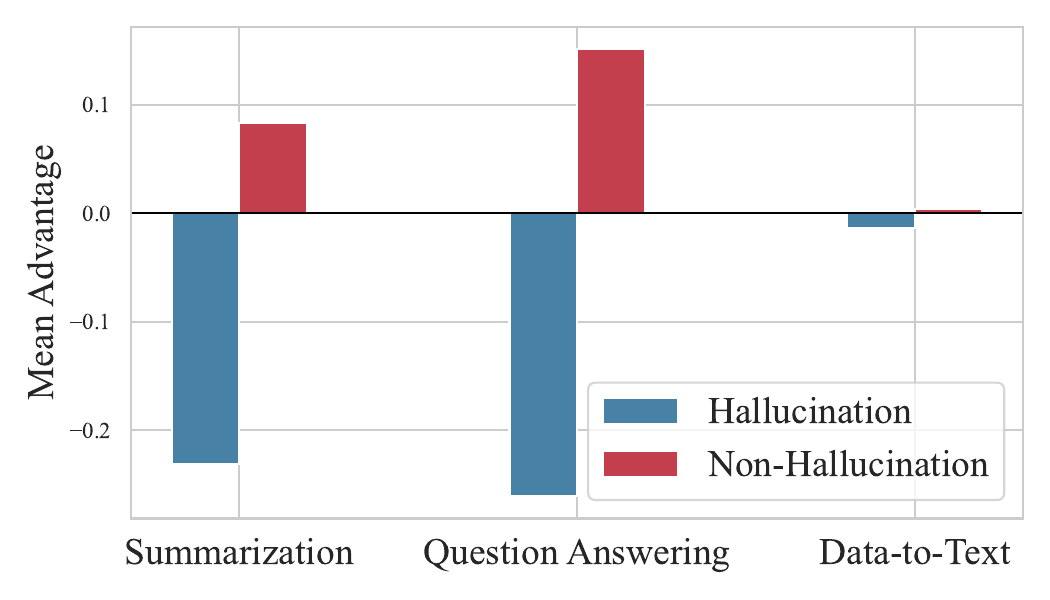}
  \caption{\textbf{Expected values of advantage given to Qwen2.5-7B-Instruct pretrained model predictions based on the prediction type.} Values are shown separately for the three task-based splits of the RAGTruth dataset.} 
  \label{fig:expectation}
  \vspace{-4em}
\end{wrapfigure}

\begin{figure*}[tp!]
    \begin{adjustbox}{width=\linewidth}
    \centering
    \includegraphics{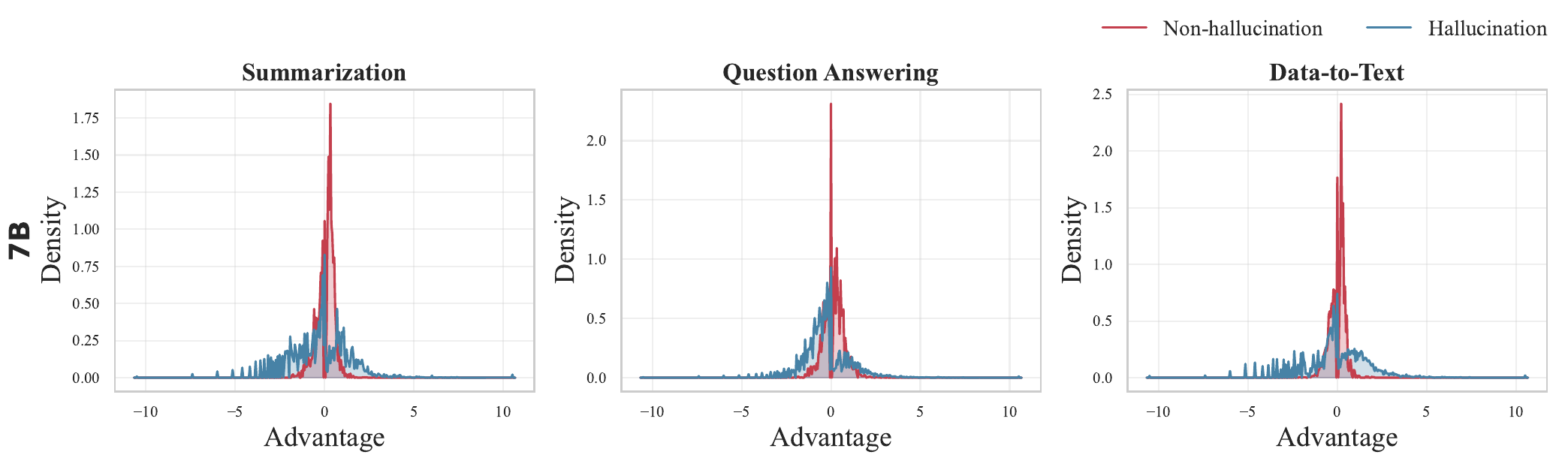}
    \end{adjustbox}
\caption{\textbf{Advantage distribution by model predictions.} 
Advantage distributions across tasks on Qwen2.5-7B-Instruct pretrained model.
Non-hallucination predictions (red) receive higher advantages than hallucination predictions (blue), revealing a class imbalance issue.}

    \label{fig:advantage_density}
\end{figure*}

Although GRPO normalizes advantages within groups, we find that the prediction type strongly biases the advantage values. As shown in Figure~\ref{fig:advantage_density}, predictions of non-hallucination consistently receive higher advantages than predictions of hallucination. Figure~\ref{fig:expectation} shows the average advantage values by prediction type confirming that predicting non-hallucination is systematically rewarded more, independent of correctness.

This stems from an inherent asymmetry in the reward function $r_{span}$. In the non-hallucination class, a model only needs to predict an empty span list to obtain a high reward. In the hallucination class, the model must precisely localize and output the correct spans. This is a harder target, and small errors cause steep drops in the F1-based reward.
As a result, GRPO tends to over-incentivize non-hallucination predictions, leading to biased behaviors with high precision but suppressed recall. 



\subsection{Class-Aware Policy Optimization}
It may seem like a natural fix to the reward asymmetry issue is to use a smaller reward value for the case $\hat{S} = S = \varnothing$. However, the standardization step used in GRPO will eliminate the effect of such scaling.  
Hence, to address this imbalance issue, we introduce Class-Aware Policy Optimization (CAPO), which uses a scaling factor $\alpha$ to scale the advantage values computed for samples that belong to the non-hallucination class.
\[
\hat{A}^{nh}_{i,t} = \alpha \cdot 
\frac{r_i - \text{mean}(\{R_j\})}{\text{std}(\{R_j\})}
\]
This formulation balances the contributions of both classes, mitigating reward sparsity in non-hallucination examples and preventing dominance by non-hallucination examples. We use $\alpha=0.5$ in our experiments. This value has been chosen based on the performance of trained model on a validation set.

\section{Experimental Setup}
We design our experiments to answer the following research questions, which structure the results and discussion (Section~\ref{sec:results}): \textbf{Q1: }What is the effectiveness of \ours?; \textbf{Q2: }Does CAPO alleviate reward hacking and achieve better precision--recall balance?; \textbf{Q3: }Is in-Domain reasoning necessary for hallucination span detection?; \textbf{Q4: }Can simply scaling rewards solve reward hacking?; \textbf{Q5: }What does \ours  learn?

\vspace{-0.5em}
\paragraph{Dataset.}  
We conduct experiments on the \textbf{RAGTruth} benchmark \cite{wu2023ragtruth} as the statistics described in Table~\ref{tab:dataset_stats}, which provides hallucination span annotations across three generation tasks: \emph{Summarization}, \emph{Question Answering (QA)}, and \emph{Data-to-Text}. Each task contains paired source documents, model-generated responses, and human-labeled hallucination spans. This makes RAGTruth one of the few datasets suitable for training and evaluating hallucination detection at the span level rather than only binary classification.
\vspace{-0.5em}
\paragraph{Models.}  
Our experiments primarily use the \texttt{Qwen2.5-7B-Instruct} and \texttt{Qwen2.5-14B-Instruct} models as base LLMs.  
For comparison, we additionally evaluate:  
\textbf{Pretrained reasoning models}: \texttt{Qwen3-8B}, \texttt{Qwen3-14B}, and \texttt{QwQ-32B}. \textbf{Proprietary reasoning models}: \texttt{GPT-5}, \texttt{o3}, \texttt{GPT-4o-mini} and \texttt{GPT-5-mini}. 
We use the default decoding strategy elaborated in the pre-trained models and use top-p = 0.95 \citep{Holtzman2020The}, top-k = 20 \citep{Holtzman2020The}, temperature = 0.6 for fine-tuned model generation.
\vspace{-0.5em}
\paragraph{Baselines.}
We compare \ours against the following approaches:  
\begin{itemize}[leftmargin=*]
    \item \textbf{Supervised Fine-Tuning (SFT)} \citep{wu2023ragtruth}: trained with cross-entropy on hallucination span annotations.  
    \item \textbf{\ours-GRPO}: our \ours approach but trained with GRPO instead of CAPO.  
    \item \textbf{Multi-View Attention} \citep{ogasa2025hallucinatedspandetectionmultiview}: token-level detector using features aggregated from multiple attention heads and attention diversity views; evaluated on attention distributions across summarization and data-to-text tasks.
\end{itemize}

\section{Results \& Discussion}
\label{sec:results}
\begin{table}[ht!]
\centering

\caption{\textbf{Span-level hallucination detection results on RAGTruth.} 
We report F1, precision, and recall across summarization, question answering, and data-to-text. 
Best scores are in bold. $^\dagger$ means the results taken from \cite{ogasa2025hallucinatedspandetectionmultiview}.
}
\begin{adjustbox}{width=\linewidth}

\begin{tabular}{ccccccccccccc}
\toprule
\multicolumn{1}{c|}{\multirow{2}{*}{\textbf{Model}}} & \multicolumn{3}{c|}{\textbf{Summarization}} & \multicolumn{3}{c|}{\textbf{Question Answering}} & \multicolumn{3}{c|}{\textbf{Data-to-Text}} & \multicolumn{3}{c}{\textbf{Avg.}} \\
\multicolumn{1}{c|}{} & F1 & Precision & \multicolumn{1}{c|}{Recall} & F1 & Precision & \multicolumn{1}{c|}{Recall} & F1 & Precision & \multicolumn{1}{c|}{Recall} & F1 & Precision & Recall \\ \midrule
\multicolumn{13}{l}{\textit{\textbf{Proprietary models}}} \\ \midrule
\multicolumn{1}{c|}{GPT-4o-mini w/ CoT} & 38.4 & 43.4 & \multicolumn{1}{c|}{34.4} & 27.3 & 33.7 & \multicolumn{1}{c|}{23.0} & 33.7 & 34.2 & \multicolumn{1}{c|}{33.2} & 33.1 & 37.1 & 30.2 \\
\multicolumn{1}{c|}{GPT-5-mini w/ CoT} & 43.9 & 33.0 & \multicolumn{1}{c|}{{65.5}} & 47.2 & 37.9 & \multicolumn{1}{c|}{62.7} & 42.5 & 29.8 & \multicolumn{1}{c|}{{74.7}} & 44.5 & 33.6 & {67.6} \\
\multicolumn{1}{c|}{GPT-5 w/ CoT} & 36.5 & 24.9 & \multicolumn{1}{c|}{68.4} & 44.4 & 32.1 & \multicolumn{1}{c|}{71.8} & 45.7 & 33.2 & \multicolumn{1}{c|}{73.5} & 42.2 & 30.0 & 71.2 \\
\multicolumn{1}{c|}{o3 w/ CoT} & 48.5 & 40.7 & \multicolumn{1}{c|}{60.1} & 49.9 & 43.4 & \multicolumn{1}{c|}{58.9} & 55.2 & 45.6 & \multicolumn{1}{c|}{70.0} & 51.2 & 43.2 & 63.0 \\ \midrule
\multicolumn{13}{l}{\textit{\textbf{Non-Reasoning models}}} \\ \midrule
\multicolumn{1}{c|}{Qwen2.5-7B-Instruct w/o CoT} & 19.3 & 28.9 & \multicolumn{1}{c|}{14.5} & 14.7 & 19.2 & \multicolumn{1}{c|}{11.9} & 14.0 & 22.3 & \multicolumn{1}{c|}{10.2} & 16.0 & 23.5 & 12.2 \\
\multicolumn{1}{c|}{Qwen2.5-7B-Instruct w/ CoT} & 21.0 & 27.4 & \multicolumn{1}{c|}{17.1} & 14.5 & 18.8 & \multicolumn{1}{c|}{11.7} & 13.0 & 32.5 & \multicolumn{1}{c|}{8.2} & 16.2 & 26.2 & 12.3 \\ 
\multicolumn{1}{c|}{Qwen2.5-14B-Instruct w/o CoT} & 31.5 & 28.0 & \multicolumn{1}{c|}{36.2} & 27.8 & 50.7 & \multicolumn{1}{c|}{55.8} & 29.0 & 22.8 & \multicolumn{1}{c|}{39.8} & 29.4 & 33.8 & 43.9 \\
\multicolumn{1}{c|}{Qwen2.5-14B-Instruct w/ CoT} & 32.9 & 44.4 & \multicolumn{1}{c|}{26.1} & 22.6 & 29.6 & \multicolumn{1}{c|}{31.6} & 26.3 & 45.0 & \multicolumn{1}{c|}{18.6} & 27.3 & 39.7 & 25.4 \\

\midrule
\multicolumn{13}{l}{\textit{\textbf{Reasoning models}}} \\ \midrule
\multicolumn{1}{c|}{QwQ-32B} & 19.4 & 50.6 & \multicolumn{1}{c|}{12.0} & 12.9 & 48.5 & \multicolumn{1}{c|}{7.5} & 13.5 & 60.7 & \multicolumn{1}{c|}{7.6} & 15.3 & 53.3 & 9.0 \\
\multicolumn{1}{c|}{Qwen3-8B} & 34.7 & 42.2 & \multicolumn{1}{c|}{29.5} & 30.5 & 32.0 & \multicolumn{1}{c|}{29.1} & 20.3 & 45.2 & \multicolumn{1}{c|}{13.1} & 28.5 & 39.8 & 23.9 \\
\multicolumn{1}{c|}{Qwen3-14B} & 35.8 & 36.9 & \multicolumn{1}{c|}{34.9} & 30.6 & 30.7 & \multicolumn{1}{c|}{30.6} & 34.8 & 40.9 & \multicolumn{1}{c|}{30.4} & 33.7 & 36.2 & 32.0 \\ \midrule
\multicolumn{13}{l}{\textit{\textbf{Finetuned models}}} \\ \midrule
\multicolumn{1}{c|}{SFT-7B} & 44.1 & 52.2 & \multicolumn{1}{c|}{38.2} & 51.3 & 51.3 & \multicolumn{1}{c|}{51.4} & 54.8 & 58.8 & \multicolumn{1}{c|}{51.5} & 50.1 & 54.1 & 47.0 \\

\multicolumn{1}{c|}{SFT-14B} & 52.7 & 57.6 & \multicolumn{1}{c|}{48.7} & 53.9 & 53.1 & \multicolumn{1}{c|}{54.8} & 59.6 & 61.6 & \multicolumn{1}{c|}{57.8} & 55.4 & 57.4 & 53.8 \\

\multicolumn{1}{c|}{Multi-View Attention-7B$^\dagger$ } & 41.5 & 49.6 & \multicolumn{1}{c|}{35.7} & 50.6 & 38.5 & \multicolumn{1}{c|}{{73.7}} & 55.2 & 53.5 & \multicolumn{1}{c|}{57.1} & 49.1 & 47.2 & 55.5 \\ \midrule
\multicolumn{13}{l}{\textit{\textbf{Ours: \ours}}} \\ \midrule
\multicolumn{1}{c|}{\ours-GRPO-7B} & 51.2 & {68.7} & \multicolumn{1}{c|}{40.9} & {55.0} & {59.6} & \multicolumn{1}{c|}{52.1} & 56.3 & 66.5 & \multicolumn{1}{c|}{48.8} & 54.2 & {64.9} & 47.3 \\
\multicolumn{1}{c|}{\ours-7B} & 50.9 & {64.4} & \multicolumn{1}{c|}{42.3} & \textbf{56.4} & {57.1} & \multicolumn{1}{c|}{56.5} & 60.4 & 67.1 & \multicolumn{1}{c|}{54.9} & 55.9 & {62.9} & 51.2 \\
\multicolumn{1}{c|}{\ours-14B} & \textbf{57.6} & 64.2 & \multicolumn{1}{c|}{52.3} & 54.8 & 52.5 & \multicolumn{1}{c|}{57.3} & \textbf{62.6} & {67.2} & \multicolumn{1}{c|}{58.7} & \textbf{58.3} & 61.3 & 56.1 \\ \bottomrule
\end{tabular}

\end{adjustbox}
\label{tab:main}
\end{table}
\subsection{\textbf{Q1: }What is the effectiveness of \ours?}
Table~\ref{tab:main} reports span-level hallucination detection results on RAGTruth across summarization, question answering, and data-to-text. 
We compare pretrained prompting baselines with models fine-tuned under our \ours framework.
\vspace{-0.5 em}
\paragraph{Pretrained instruction-tuned models.} Qwen2.5-7B/14B-Instruct, with or without CoT, perform poorly (F1 below 30), indicating that prompting alone is insufficient for accurate span localization.
\vspace{-0.5 em}

\paragraph{Pretrained reasoning models.} Models designed for reasoning (QwQ-32B, Qwen3-8B, Qwen3-14B) transfer some reasoning ability to hallucination detection. For example, Qwen3-14B improves summarization F1 to 35.8 compared to 32.9 for Qwen2.5-14B-Instruct. However, these models still trail fine-tuned approaches, showing that general reasoning ability alone is insufficient for span-level detection.
\vspace{-0.5 em}
\paragraph{Finetuned baselines.} Supervised fine-tuning (SFT) provides strong gains, reaching 55.4 F1 at 14B scale. Multi-View Attention~\citep{ogasa2025hallucinatedspandetectionmultiview} further pushes the 7B model to 49.1 F1, though still behind larger SFT models.
\vspace{-0.5 em}
\paragraph{\ours} \ours consistently outperforms all baselines, including proprietary GPT-4o/5-mini, GPT-5, and o3. \ours-7B outperforms SFT on all three tasks (avg. 55.9 v.s 50.1). At 14B, \ours-14B achieves 57.6 on summarization, 54.8 on QA, and 62.6 on Data-to-Text, surpassing Qwen3 and the strongest GPT-5 and o3 models. This establishes \ours demonstrating that reinforcement learning with span-level rewards effectively aligns reasoning with hallucination detection.

\subsection{\textbf{Q2: }Does CAPO alleviate reward hacking and achieve better precision--recall balance?}
\begin{figure*}[tp!]
    \begin{adjustbox}{width=\linewidth}
    \centering
    \includegraphics{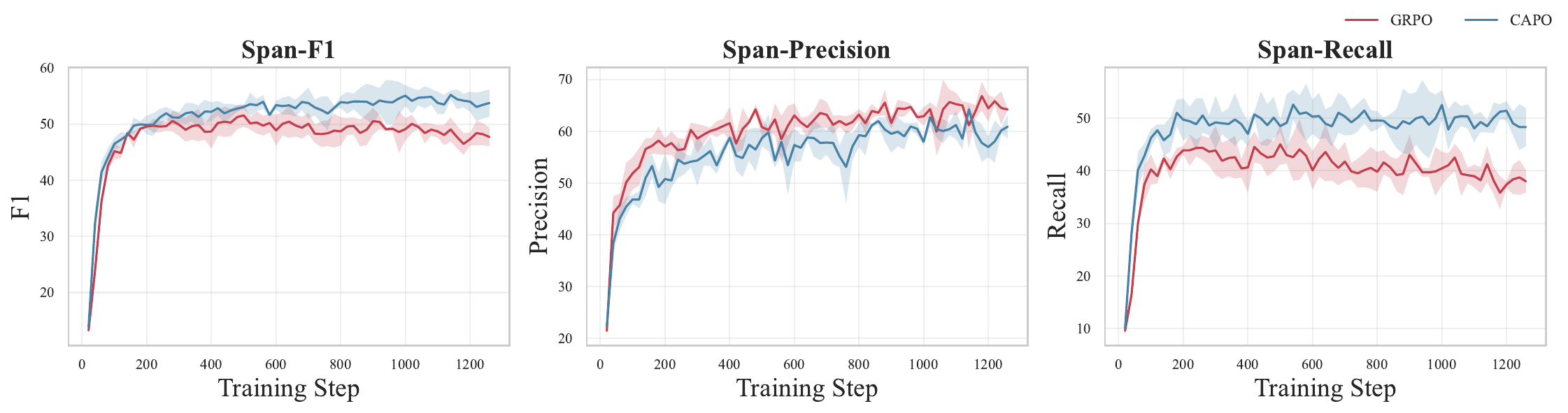}
    \end{adjustbox}
\caption{\textbf{Training dynamics of GRPO (red) and CAPO (blue) on Qwen2.5-7B-Instruct model.}  
While GRPO exhibits high precision but declining recall due to reward hacking, CAPO stabilizes recall without sacrificing precision, yielding consistently higher span F1. Shaded regions denote standard deviations across runs.}

     \label{fig:val_curve}
\end{figure*}
A key limitation we observed with GRPO is that models often exploit the reward design by defaulting to predicting no hallucination spans, which yields high precision but severely hurts recall. This behavior reflects a form of \emph{reward hacking}, where the model learns shortcuts that maximize rewards without genuinely improving hallucination detection. As shown in our advantage distribution analysis (Figure~\ref{fig:advantage_density}), predictions of non-hallucination systematically receive higher advantages, biasing the policy toward conservative behavior.

Figure~\ref{fig:val_curve} compares training dynamics of GRPO and our proposed CAPO across span F1, precision, and recall. We make two observations: (1) \textbf{GRPO favors precision over recall.} As training progresses, GRPO maintains relatively high precision but recall gradually drops, showing the model’s tendency to avoid making positive span predictions.; (2) \textbf{CAPO balances precision and recall.} By re-weighting class-specific advantages, CAPO stabilizes recall while preserving strong precision, resulting in a clear improvement in span F1 throughout training.


These results confirm that CAPO directly addresses the imbalance highlighted in our advantage distribution analysis. By correcting for class-dependent reward sparsity, CAPO mitigates reward hacking and achieves a better precision--recall trade-off, consistently yielding higher span F1 compared to vanilla GRPO.

\subsection{\textbf{Q3: }Is in-Domain reasoning necessary for hallucination span detection?} 
\begin{figure*}[tp!]

    \begin{adjustbox}{width=\linewidth}
    \centering
    \includegraphics{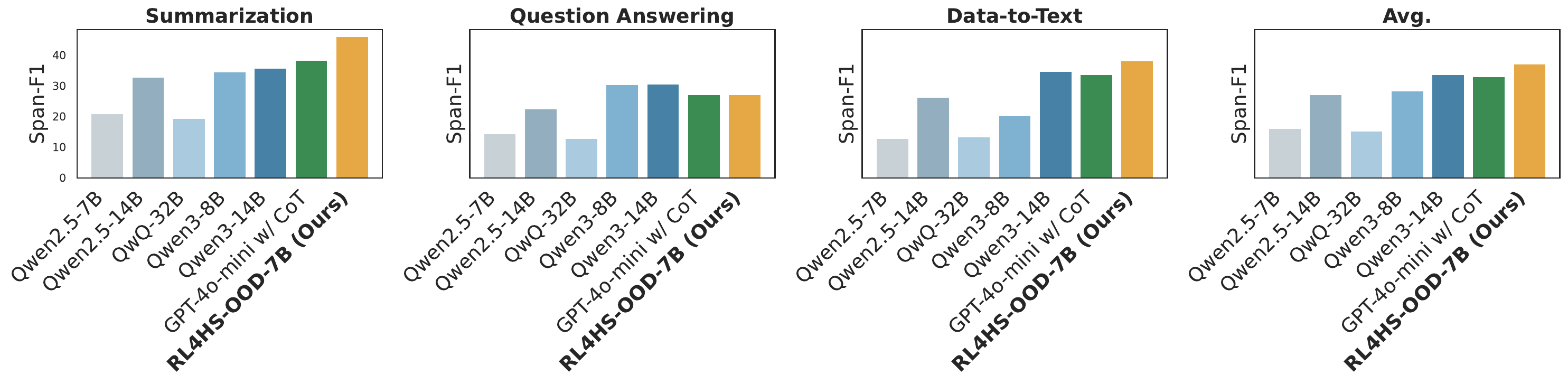}
    \end{adjustbox}
\caption{\textbf{Out-of-domain evaluation on RAGTruth.} Span-F1 scores on Ragtruth dataset. Our \ours-OOD-7B model performs competitively with larger reasoning models, showing the benefit of span-level reward fine-tuning. We use Instruct models for Qwen2.5 models.}

     \label{fig:ood_bar}
\end{figure*}

To assess whether hallucination span detection requires in-domain reasoning rather than generic reasoning ability, we conduct leave-one-out training with \ours (\ours-OOD-7B), holding out one task at a time and evaluating on the unseen task. Figure~\ref{fig:ood_bar} shows results compared against reasoning-focused models (QwQ, Qwen3 and GPT-4o-mini with COT).

General-purpose reasoning models such as Qwen3 and QwQ transfer some reasoning ability but their Span-F1 scores often remain below 40, showing that generic reasoning is insufficient for fine-grained hallucination detection. RL4HS-OOD-7B, in contrast, achieves consistently stronger results across all held-out tasks. Moreover, despite being much smaller, RL4HS-OOD-7B performs better than GPT-4-mini with COT, underscoring the efficiency of span-level reward fine-tuning. These findings highlight that in-domain "reasoning" learned with span-level rewards is essential for robust hallucination detection.

\vspace{10pt}

\subsection{\textbf{Q4: }Can simply scaling rewards solve reward hacking?}
\begin{wraptable}{r}{0.5\textwidth}
\centering
\caption{\textbf{Comparison of GRPO, CAPO, and Dr.GRPO variants with \ours.} CAPO improves F1 by addressing reward imbalance, while Dr.GRPO with different $\gamma$ values shows varying precision–recall trade-offs but does not surpass CAPO.}
\small
\begin{tabular}{c|ccc}

\toprule
\multirow{2}{*}{\textbf{Method}} & \multicolumn{3}{c}{\textbf{Avg.}} \\
 & F1 & Precision & Recall \\ \midrule
GRPO & 54.2 & 64.9 & 47.3 \\
CAPO & \textbf{55.9} & 62.9 & {51.2} \\ \midrule
Dr.GRPO w/ $\gamma$=0.1 & 52.5 & 53.6 & 52.3 \\
Dr.GRPO w/ $\gamma$=0.5 & 54.7 & 62.2 & 49.4 \\
Dr.GRPO w/ $\gamma$=1.0 & {53.1} & {64.1} & 45.8 \\ \bottomrule
\end{tabular}

\label{tab:dr_grpo}
\end{wraptable}

One concern with GRPO is that its standardization of group rewards diminishes the intrinsic difficulty difference between hallucination and non-hallucination cases, often biasing the model toward predicting non-hallucination. To address this, we explored a variant of Dr.GRPO \citep{liu2025understandingr1zeroliketrainingcritical}, which removes standardization and instead scales the reward for successfully predict non-hallucination by a factor $\gamma$
(i.e. $r_{\text{span}} = \gamma, \quad \text{if } \hat{S} = \varnothing \text{ and } S = \varnothing$).
Table~\ref{tab:dr_grpo} reports results under different $\gamma$ values. While Dr.GRPO influences the precision–recall tradeoff (e.g., higher $\gamma$ increases recall at the cost of precision), overall performance is inferior to standard GRPO and \ours. This suggests that the normalization step in GRPO is crucial, and simple reward rescaling cannot effectively address reward hacking in our task.


%



\subsection{\textbf{Q5: }What does \ours learn? A case study}
To better understand the reasoning behaviors learned by \ours, we examine qualitative outputs on the RAGTruth dataset (Table~\ref{tab:example}). The example highlights a discrepancy regarding whether the restaurant provides catering services. 
\textbf{Pretrained model.} Before training, the pretrained model fails to identify the inconsistency. Although it checks structured business hours and customer reviews, it overlooks the fact that the structured data contains no attribute related to catering services. As a result, the model produces no hallucination spans.
\textbf{\ours.} In contrast, \ours correctly flags the catering services claim as a hallucination. Its reasoning process closely mirrors the human-designed heuristic pipeline:  
\begin{itemize}[leftmargin=*]
    \item Step 1: Identify explicit claims in the article (e.g., “provides catering services”).  
    \item Step 2: Cross-check these claims against structured business data (which does not list catering services as an attribute).  
    \item Step 3: Conclude that the claim is inconsistent and mark it as hallucinated.  
\end{itemize}

This case demonstrates that \ours goes beyond surface-level reasoning traces. Instead of producing generic or irrelevant explanations, the model performs systematic consistency checks that align with heuristic rules used in prior hallucination detection pipelines. This suggests that the reasoning behavior learned under span-level rewards is genuine, faithful, and semantically grounded.

\begin{table}[ht!]

\caption{\textbf{Case study comparing pretrained and \ours models on detecting hallucinations.}}
\centering
\begin{adjustbox}{width=\linewidth}
\begin{tabular}{l}
\toprule
\textit{\textbf{Review data}} \\ \midrule
\begin{tabular}[c]{@{}l@{}}’name’: ’Benchmark Eatery’, ’address’: ’1201 State St’, ’city’: ’Santa Barbara’, ’state’: ’CA’, ’categories’: ’American (Traditional), American\\ (New), Breakfast \& Brunch, Restaurants, Seafood, Vegetarian, Nightlife, Event Planning \& Services, Bars, Venues \& Event Spaces’, ’hours’: ’Monday’:\\ ’0:0 0:0’, ’Tuesday’: ’11:30-20:0’, ’Wednesday’: ’11:30-20:0’, ’Thursday’: ’11:30-20:0’, ’Friday’: ’11:30-16:0’, ’Saturday’: ’11:30-16:0’, ’Sunday’:\\ ’11:30-20:0’, ’attributes’: ’BusinessParking’: ’garage’: True, ’street’: True, ’validated’: False, ’lot’: True, ’valet’: False, ’RestaurantsReservations’:\\ False, ’OutdoorSeating’: True, ’WiFi’: ’free’, ’RestaurantsTakeOut’: True, ’RestaurantsGoodForGroups’: True, ’Music’: False, ’Ambience’: ’touristy’:\\ False, ’hipster’: False, ’romantic’: False, ’divey’: False, ’intimate’: False, ’trendy’: False, ’upscale’: False, ’classy’: True, ’casual’: True, ’business stars’:\\ 4.0, ’review info’: {[}’review stars’: 4.0, ’review date’: ’2022-01-02 21:02:49’, ’review text’: ’Nice little place in downtown Santa Barbara where we\\ stopped for late lunch/early dinner on our way back home to San Diego. We loved our flatbreads. I had been craving clam chowder and ordered it here.\\ It was a little thick for my liking, but tasted good. Nice, friendly upscale casual place on State Street.’, ’review stars’: 1.0, ’review date’: ’2021-12-20\\ 22:29:32’, ’review text’: ”I had high hopes for this place, but it fell flat big time. I ordered the BLAT with grilled chicken and a side salad. I specifically\\ asked for no tomato and they put tomato on both the sandwich and the salad. Not a huge deal, just annoying to pick it all off. And if you have an allergy,\\ I would not trust this place to be mindful of that. The grilled chicken was not seasoned at all and the food overall was just extremely bland. My sandwich\\ was stacked so high that the avocado all fell out as I was eating it. The side salad is lettuce, cucumber, red onion, and tomato. My bf got a burger with\\ onion rings and he did not enjoy either part of the meal. The onion rings are beer battered and basically all batter. The burger was also unseasoned and\\ not enjoyable. The service is ok. The staff is nice enough but my bf did have to get up and ask for his soda to be refilled and we were never offered\\ a refill on our waters. We both left the restaurant with a stomach ache, I personally felt ill. I had stomach pains for a few hours afterwards and I can\\ only assume it’s because the food is not fresh because my meal didn’t have any allergens that would’ve given me that reaction. Lastly, the cleaning spray\\ they used on the tables just smelled funky. Idk how to explain it but we actually moved to an outside table because the inside one smelled so musty and\\ mildewy.patio is nice, but not worth coming back for. Additionally, the menu is too scattered and overwhelming. I think they should stick to one type\\ of food and work on increasing the quality rather than having too many different menu options that aren’t cohesive.”, ’review stars’: 5.0, ’review date’:\\ ’2021-11-29 20:45:43’, ’review text’: ’We have been here a few times during trial over the past couple weeks and it is awesome. I keep ordering the\\ burrito but everyone seems to enjoy their food. The one piece of feedback would be the Spanish soup seems to change - from not too spicy to spicy.’{]}\end{tabular} \\ \midrule
\textit{\textbf{Machine-generated article given review data}} \\ \midrule
\begin{tabular}[c]{@{}l@{}}Benchmark Eatery is an American-style restaurant located in downtown Santa\\ Barbara, California. The restaurant offers a variety of cuisines, including traditional American, seafood, vegetarian, and event planning services. They also\\ provide outdoor seating, free Wi-Fi, takeout options, and \colorbox{AppleRed2}{catering services}. According to customer reviews, the restaurant has a four-star rating, with mixed\\ opinions about the quality of the food and service. Some customers have praised the restaurant’s flatbreads and clam chowder, while others have criticized\\ the restaurant for poor flavor, blandness, and inconsistent ingredients. Additionally, some customers have expressed concerns about the restaurant’s\\ handling of allergies and the cleanliness of the establishment. Overall, Benchmark Eatery appears to be a popular choice for dining in downtown\\ Santa Barbara, with a range of menu options and a cozy atmosphere.\end{tabular} \\ \midrule
\textit{\textbf{Before \ours-7B (Pretrained model)}} \\ \midrule
\begin{tabular}[c]{@{}l@{}}\#\# Step 1: \colorbox{AppleRed3}{Analyze the structured data to understand the restaurant’s categories and services}.\\ The restaurant is categorized as offering American (Traditional) and American (New) cuisine, as well as Breakfast \& Brunch. It also provides seafood,\\ vegetarian options, event planning services, outdoor seating, free Wi-Fi, takeout options, and catering services. \#\# Step 2: Compare the information in the\\ article with the structured data. The article correctly mentions that Benchmark Eatery is an American-style restaurant located in downtown Santa Barbara.\\ It also correctly states that they offer a variety of cuisines including traditional American and seafood. The article accurately notes that they provide\\ outdoor seating and free Wi-Fi. \#\# Step 3: Check for any discrepancies between the review information provided in structured data and mentioned in the\\ article. The review information includes both positive feedback about flatbreads and clam chowder from some customers, as well as negative feedback\\ about poor flavor, blandness from others. The article accurately captures these mixed opinions. \#\# Step 4: Verify if there are any inconsistencies regarding\\ food handling practices or cleanliness. The structured data does not explicitly mention concerns about handling allergies or cleanliness issues; however,\\ one customer review did express concern about inconsistent ingredients used in their meal which could imply some issues with food quality or preparation\\ consistency. The article mentions concerns about handling allergies but does not specifically state cleanliness issues based on this data alone. \#\# Step 5:\\ Confirm overall rating based on customer reviews. The business has a four-star rating according to customer reviews mentioned in both sources. Based\\ on this analysis: \colorbox{AppleRed3}{\textbf{\{”hallucination list”: {[}{]}\}}}\end{tabular} \\ \midrule
\textit{\textbf{\ours-7B}} \\ \midrule
\begin{tabular}[c]{@{}l@{}}\#\# Step 1: \colorbox{AppleBlue2}{Check for Catering Services} The article states ”They also provide outdoor seating, free Wi-Fi, takeout options, and catering services.” However,\\ per data, catering services seems inconsistent. It seems to say ”catering services” which seems inconsistent. It appears to be an error; suspecting potential\\ candidates of typo or misinterpretation. \#\# Step 2: \colorbox{AppleBlue2}{Evaluate Food Quality and Services.} The text says that according to customer’s opinion’s the restaurant\\ has a four-star rating with mixed food quality and service as discussed. \#\# Step 3: \colorbox{AppleBlue2}{Validate Food Quality’s.} It states that ”Some customers have praised\\ the restaurant’s flatbreads and clam chowder, while others have criticized the restaurant for poor flavor, blandness, and inconsistent ingredients.” This\\ part is confirmed by our data. \#\# Step 4: \colorbox{AppleBlue2}{Evaluate Customer’s Concerns.} It says customers expressed concerns about food quality again as well as\\ cleanliness of establishment; this part seems valid per our data. \#\# Step 5: \colorbox{AppleBlue2}{Check for Food Variety.} It describes variety of cuisines like traditional\\ American (both traditional \& new), breakfast \& brunch, seafood, vegetarian food etc. again as said in text. \#\# Step 6: \colorbox{AppleBlue2}{Evaluate Overall Assessment}\\ of Restaurant’s Attraction. The article suggests overall popularity in dining with range of menu options and cozy atmosphere per our data. Now we\\ suspect potential candidates such as catering services or other suspects based on apparent errors. If it says catering services it seems suspicious. The\\ candidate suspects are often candidates of typographical errors or misinterpretation: \#\#\# Compiled Results: \colorbox{AppleBlue2}{\textbf{\{”hallucination list”: {[}”catering services”{]}\}}}\end{tabular} \\ \bottomrule
\end{tabular}

\end{adjustbox}
\label{tab:example}
\end{table}

\section{Related Works}
\paragraph{Hallucination Detection.}
Hallucination detection research has evolved from binary classification to fine-grained span detection. Early work focused on binary judgments—whether text contains hallucinations \citep{manakul-etal-2023-selfcheckgpt, luo2023chatgptfactualinconsistencyevaluator, tang-etal-2024-minicheck}. However these approach failed to localize where the hallucination. 
\cite{yang-etal-2024-fizz, scire-etal-2024-fenice} proposed a cascade pipeline that leverage atomic-fact generation, natural language inference to detection hallucination. But the pipeline is hard to optimize.
Recent methods target span-level detection.  introduced RAGTruth \citep{wu2023ragtruth} with human-annotated spans across three generation tasks. \cite{ogasa2025hallucinatedspandetectionmultiview} aggregated multi-head attention features for token-level detection. However, these attention-based methods lack explicit reasoning mechanisms. 
\paragraph{Reasoning Enhancement in NLP.}
 Group Relative Policy Optimization (GRPO), originally developed to improve mathematical reasoning by comparing groups of outputs rather than relying on a separate value model.
 GRPO has since been extended and adapted to a variety tasks such coding \citep{code-r1, chen2025r1codeinterpretertrainingllmsreason}, planning \citep{hao2023reasoninglanguagemodelplanning}, tool-calling \citep{feng2025retoolreinforcementlearningstrategic, shang2025rstar2agentagenticreasoningtechnical}.
 More recently, researchers has show that GRPO can also be applied to enhance reasoning in traditional NLP tasks such as NLI \citep{shao2024deepseekmathpushinglimitsmathematical}, intent classification \citep{feng2025improvinggeneralizationintentdetection}, and safety alignment \cite{li2025optimizingsafealignedlanguage}. Showing the effectiveness of GRPO with LLM.

\section{Conclusion}
We introduced \ours, a reinforcement learning framework that uses span-level rewards to align LLM reasoning with hallucination detection. While CoT offers limited single-sample gains, \ours distills its multi-sample advantages into stronger predictions. With CAPO to address reward imbalance, \ours outperforms pretrained reasoning models and SFT on RAGTruth, and produces faithful, heuristic-like reasoning traces that improve both accuracy and robustness.
\section{Acknowledgments}
This work was conducted during an internship at Apple AIML. We sincerely thank Leon Gatys, Bo-Hsiang (Andy) Tseng, Han-Byul Kim and Fartash Faghri for their valuable feedback and insightful suggestions on this work. 
\bibliographystyle{plainnat}

\bibliography{Apple}

\appendix
\section{Appendix}
\subsection{Prompt}
\begin{tcolorbox}[
  colback=white,      
  colframe=AppleBlack,     
  coltitle=white,     
  title=\textbf{COT for Summarization}
]
"Below is the original document:"

\{reference\}

"Below is a summary of the document:"

\{response\}

"Your task is to determine whether the summary contains hallucinations."
"First, provide reasoning with the following format:"

\#\# Step 1: $<$ your first reasoning step $>$

\#\# Step 2: $<$ your next reasoning step $>$

...(add as many steps as needed)
Then, compile the labeled hallucinated spans into a JSON dict, with a key hallucination list and its value is a list of hallucinated spans. If there are potential hallucinations, the output should be in the following JSON format: {hallucination list: [hallucination span1, hallucination span2, ...]}. Otherwise, leave the value as an empty list as follows: {hallucination list: []}.
\end{tcolorbox}
\begin{tcolorbox}[
  colback=white,      
  colframe=AppleBlack,     
  coltitle=white,     
  title=\textbf{COT for Question Answering}
]
"Below is a question:"

\{question\}

"Below are the related passages:"

\{reference\}

"Below is an answer:"

\{response\}

"Your task is to determine whether the answer contains hallucinations."
"First, provide reasoning with the following format:"

\#\# Step 1: $<$ your first reasoning step $>$

\#\# Step 2: $<$ your next reasoning step $>$

...(add as many steps as needed)
Then, compile the labeled hallucinated spans into a JSON dict, with a key hallucination list and its value is a list of hallucinated spans. If there are potential hallucinations, the output should be in the following JSON format: {hallucination list: [hallucination span1, hallucination span2, ...]}. Otherwise, leave the value as an empty list as follows: {hallucination list: []}.
\end{tcolorbox}
\begin{tcolorbox}[
  colback=white,      
  colframe=AppleBlack,     
  coltitle=white,     
  title=\textbf{COT for Data-to-text}
]
"Below is structured data in JSON format:"

\{reference\}

Below is an overview article written in accordance with the structured data:"

\{response\}

"Your task is to determine whether the article contains hallucinations."
"First, provide reasoning with the following format:"

\#\# Step 1: $<$ your first reasoning step $>$

\#\# Step 2: $<$ your next reasoning step $>$

...(add as many steps as needed)
Then, compile the labeled hallucinated spans into a JSON dict, with a key hallucination list and its value is a list of hallucinated spans. If there are potential hallucinations, the output should be in the following JSON format: {hallucination list: [hallucination span1, hallucination span2, ...]}. Otherwise, leave the value as an empty list as follows: {hallucination list: []}.
\end{tcolorbox}


\begin{tcolorbox}[
  colback=white,      
  colframe=AppleBlack,     
  coltitle=white,     
  title=\textbf{w/o COT for Summarization}
]
"Below is the original document:"

\{reference\}

"Below is a summary of the document:"

\{response\}

"Your task is to determine whether the summary contains hallucinations."

Then, compile the labeled hallucinated spans into a JSON dict, with a key hallucination list and its value is a list of hallucinated spans. If there are potential hallucinations, the output should be in the following JSON format: {hallucination list: [hallucination span1, hallucination span2, ...]}. Otherwise, leave the value as an empty list as follows: {hallucination list: []}.
\end{tcolorbox}
\begin{tcolorbox}[
  colback=white,      
  colframe=AppleBlack,     
  coltitle=white,     
  title=\textbf{w/o COT for Question Answering}
]
"Below is a question:"

\{question\}

"Below are the related passages:"

\{reference\}

"Below is an answer:"

\{response\}

"Your task is to determine whether the answer contains hallucinations."

Then, compile the labeled hallucinated spans into a JSON dict, with a key hallucination list and its value is a list of hallucinated spans. If there are potential hallucinations, the output should be in the following JSON format: {hallucination list: [hallucination span1, hallucination span2, ...]}. Otherwise, leave the value as an empty list as follows: {hallucination list: []}.
\end{tcolorbox}
\begin{tcolorbox}[
  colback=white,      
  colframe=AppleBlack,     
  coltitle=white,     
  title=\textbf{w/o COT for Data-to-text}
]
"Below is structured data in JSON format:"

\{reference\}

Below is an overview article written in accordance with the structured data:"

\{response\}

"Your task is to determine whether the article contains hallucinations."

Then, compile the labeled hallucinated spans into a JSON dict, with a key hallucination list and its value is a list of hallucinated spans. If there are potential hallucinations, the output should be in the following JSON format: {hallucination list: [hallucination span1, hallucination span2, ...]}. Otherwise, leave the value as an empty list as follows: {hallucination list: []}.
\end{tcolorbox}
\subsection{Training Details}
\begin{table}[h]

\caption{\textbf{Training details for SFT and RL.}}
\centering
\begin{adjustbox}{width=0.5\linewidth}
\begin{tabular}{cccc}
\toprule
Method & Size & Learning Rate & Batch Size \\ \midrule
\multirow{2}{*}{SFT} & 7B & 1e-6 & 64 \\
 & 14B & 1e-6 & 64 \\ \midrule
\multirow{2}{*}{RL} & 7B & 1e-6 & 64 \\
 & 14B & 5e-7 & 64 \\ \bottomrule
\end{tabular}


\end{adjustbox}

\label{tab:details}
\end{table}
We trained our models using 8 H100 GPUs. The learning rate and batch size configurations are provided in Table~\ref{tab:details}. For reinforcement learning training, we set the group size to 16 and used rollout generation with temperature = 1.0, top-p = 1.0, and top-k = -1. Following \cite{yu2025dapoopensourcellmreinforcement}, we also adopted a clipping threshold of clip\_high = 0.28. Due to the lack of the reasoning data, we fine-tuned instruct model with RL directly instead of doing SFT first.

For GPT-series models, we used top-p = 0.95 and temperature = 0.7 to generate response during inference.
All the trained models were trained with 5 epochs and selected the checkpoints with the best performance on self-splitted validation set.
In our training, we resolved the data class imbalance by upweighting hallucination class to have equal amount of data. 


\subsection{Dataset Statistic}
\FloatBarrier
\begin{table}[h!]
\caption{
\textbf{Dataset statistics for RAGTruth.} Numbers indicate the number of hallucination examples, with the number of non hallucination examples shown in parentheses.
}

\centering
\begin{tabular}{lccc}
\toprule
& \textbf{Summarization} & \textbf{Question Answering} & \textbf{Data-to-Text} \\
\midrule
Train & 1209 (2646) & 1277 (2732) & 3048 (1347) \\
Val   & 271 (629)  & 269 (614)   & 624 (276)   \\
Test  & 204 (696)  & 160 (715)   & 579 (321)   \\
\bottomrule
\end{tabular}
\label{tab:dataset_stats}
\end{table}
\FloatBarrier

\subsection{F1@K}

\FloatBarrier
\label{sec:f1@kapp}
\begin{figure*}[h!]
    \begin{adjustbox}{width=\linewidth}
    \centering
    \includegraphics{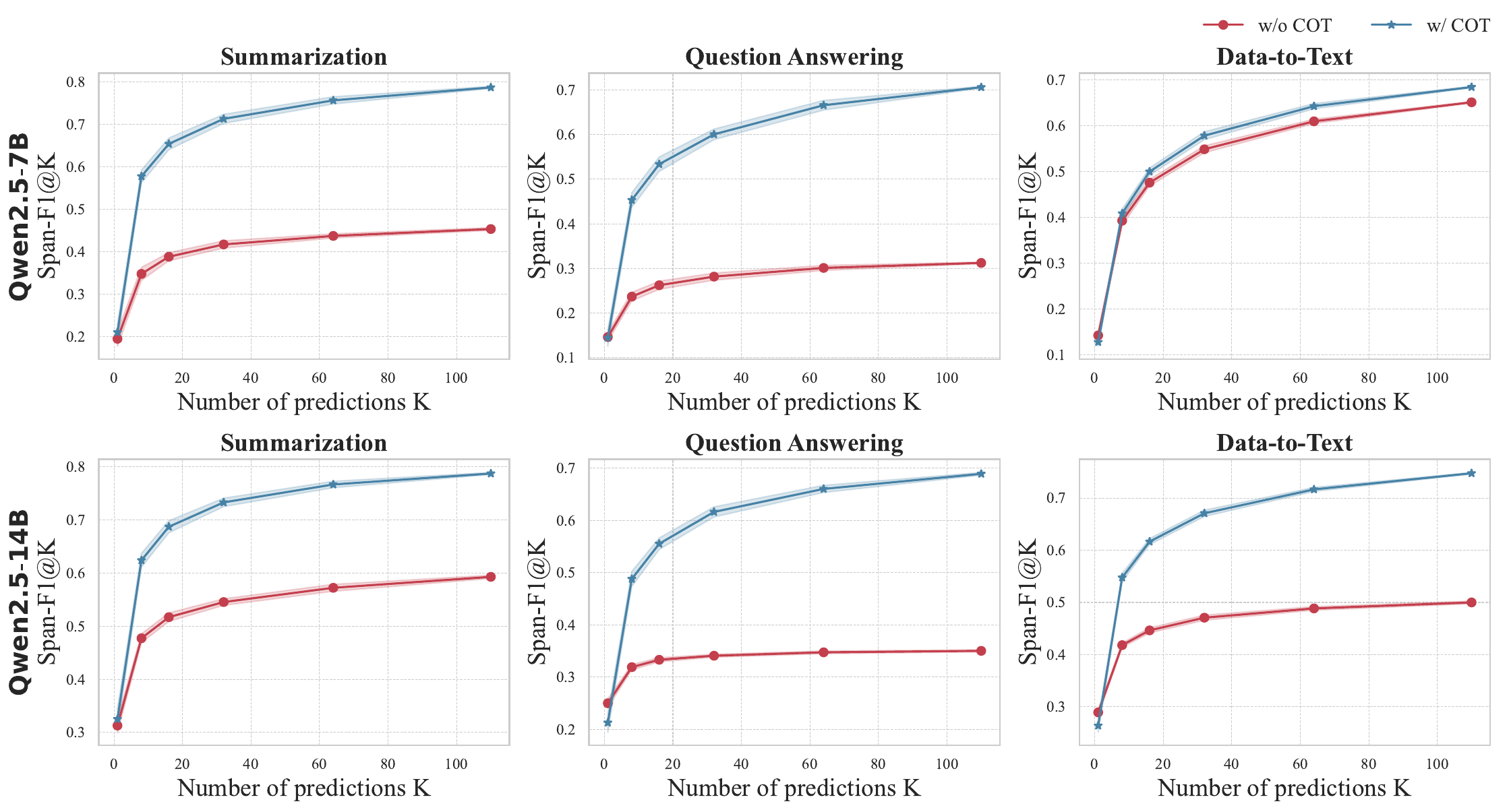}
    \end{adjustbox}
    \caption{Hallucination span detection with and without CoT reasoning.   
    Results are shown for summarization, question answering, and data-to-text tasks on the RAGTruth benchmark.}  
     \label{fig:f1@k_qwen2}
\end{figure*}

\begin{figure*}[h!]
    \begin{adjustbox}{width=\linewidth}
    \centering
    \includegraphics{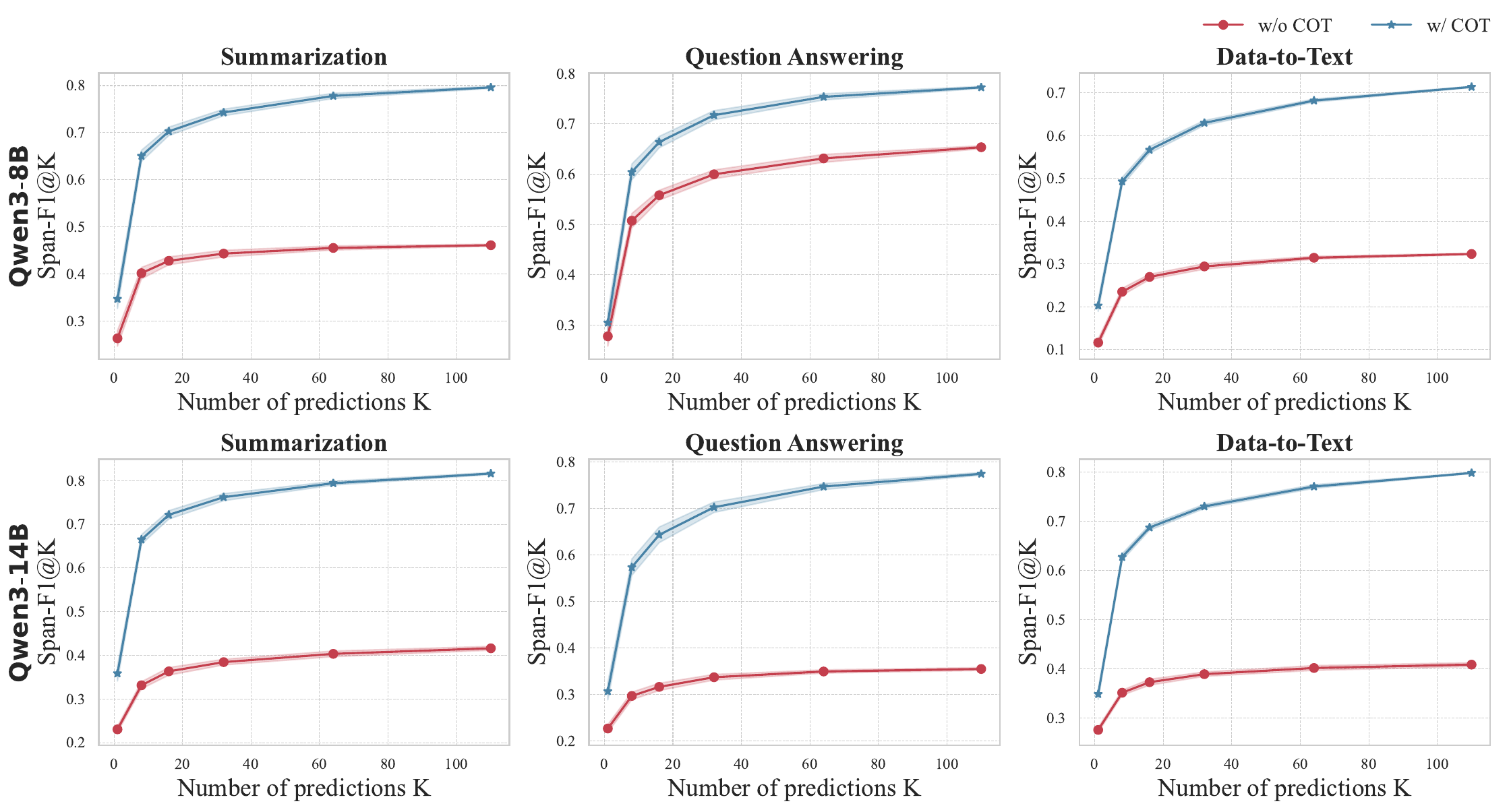}
    \end{adjustbox}
    \caption{Hallucination span detection with and without CoT reasoning.   
    Results are shown for summarization, question answering, and data-to-text tasks on the RAGTruth benchmark.}  
     \label{fig:f1@k_qwen3}
\end{figure*}


\FloatBarrier

\applefootnote{ \textcolor{textgray}{\sffamily Apple and the Apple logo are trademarks of Apple Inc., registered in the U.S. and other countries and regions.}}

\end{document}